\documentclass[runningheads]{llncs}
\usepackage[dvipdfmx]{graphicx}
\usepackage{amsmath, amssymb}

\DeclareMathOperator*{\argmin}{argmin}

\usepackage{booktabs}
\usepackage{multirow}
\usepackage{color, colortbl}
\definecolor{LightGray}{rgb}{0.97,0.97,0.97}

\usepackage[whole]{bxcjkjatype}

\begin{document}
\title{Reducing the Cost: Cross-Prompt Pre-Finetuning for Short Answer Scoring}
%
%
\author{Hiroaki Funayama\inst{1,2}, Yuya Asazuma\inst{1,2}, Yuichiroh Matsubayashi\inst{1,2}, \\  Tomoya Mizumoto\inst{2} and Kentaro Inui\inst{1,2}
}
\authorrunning{Funayama et al.}
%
\institute{
 Tohoku University, Sendai, Japan\\
 \email{\{h.funa, 	asazuma.yuya.r7\}@dc.tohoku.ac.jp,} \email{\{y.m, 	inui\}@tohoku.ac.jp,}
 \and
 RIKEN, Tokyo, Japan\\
 \email{tomoya.mizumoto@a.riken.jp}
 }
\maketitle              
\begin{abstract}

Automated Short Answer Scoring (SAS) is the task of automatically scoring a given input to a prompt based on rubrics and reference answers.
Although SAS is useful in real-world applications, both rubrics and reference answers differ between prompts, thus requiring a need to acquire new data and train a model for each new prompt.
Such requirements are costly, especially for schools and online courses where resources are limited and only a few prompts are used. 
In this work, we attempt to reduce this cost through a two-phase approach: train a model on existing rubrics and answers with gold score signals and finetune it on a new prompt.
Specifically, given that scoring rubrics and reference answers differ for each prompt, we utilize key phrases, or representative expressions that the answer should contain to increase scores, and train a SAS model to learn the relationship between key phrases and answers using already annotated prompts (i.e., cross-prompts).
Our experimental results show that finetuning on existing cross-prompt data with key phrases significantly improves scoring accuracy, especially when the training data is limited.
Finally, our extensive analysis shows that it is crucial to design the model so that it can learn the task's general property.
We publicly release our code and all of the experimental settings for reproducing our results \footnote{\url{https://github.com/hiro819/Reducing-the-cost-cross-prompt-prefinetuning-for-SAS.git}}.

\end{abstract}

\section{Introduction}
\label{sec:introduction}
Automated Short Answer Scoring (SAS) is the task of automatically scoring a given student answer to a prompt based on existing rubrics and reference answers~\cite{Leacock,Mohler:11:ACL,sultan-etal-2016-fast,Surya-deep-learning-shortanswer-2019}. 
SAS has been extensively studied as a means to reduce the burden of manually scoring student answers in school education and large-scale examinations or as a technology for augmenting e-learning environments~\cite{Singla2019GetIS,roy-etal-2016-wisdom,Zhai2021-practices}.
However, SAS in the practical application requires one critical issue to be addressed: the cost of preparing training data. 
Data to train SAS models (i.e. students answers with human-annotated gold score signals) must be prepared for each prompt independently, as the rubrics and reference answers are different for each prompt~\cite{Burrows2015}.

In this paper, we address this issue by exploring the potential benefit of using the \textit{cross-prompt} training data, or training data consisting of different prompts, in model training. 
The cost of preparing training data will be alleviated if a SAS model can leverage cross-prompt data to boost the scoring performance with the same amount of in-prompt data. 
However, this approach imposes two challenges. 
First, it is not obvious whether a model can learn from cross-prompt data something useful for scoring answers to a new target prompt, as the new prompt must have different scoring criteria from cross-prompt data available a priori (\emph{cross-prompt generalizability}). 
Second, in a real-world setting, cross-prompt data (possibly proprietary) may not be accessible when classrooms or e-learning courses train a new model for their new prompts (\emph{data accessibility}). 
Therefore, we want an approach where one can train a model for a new prompt without accessing cross-prompt data while benefitting from cross-prompt training. 

\begin{figure}[t]
\centering
\includegraphics[width=10cm]{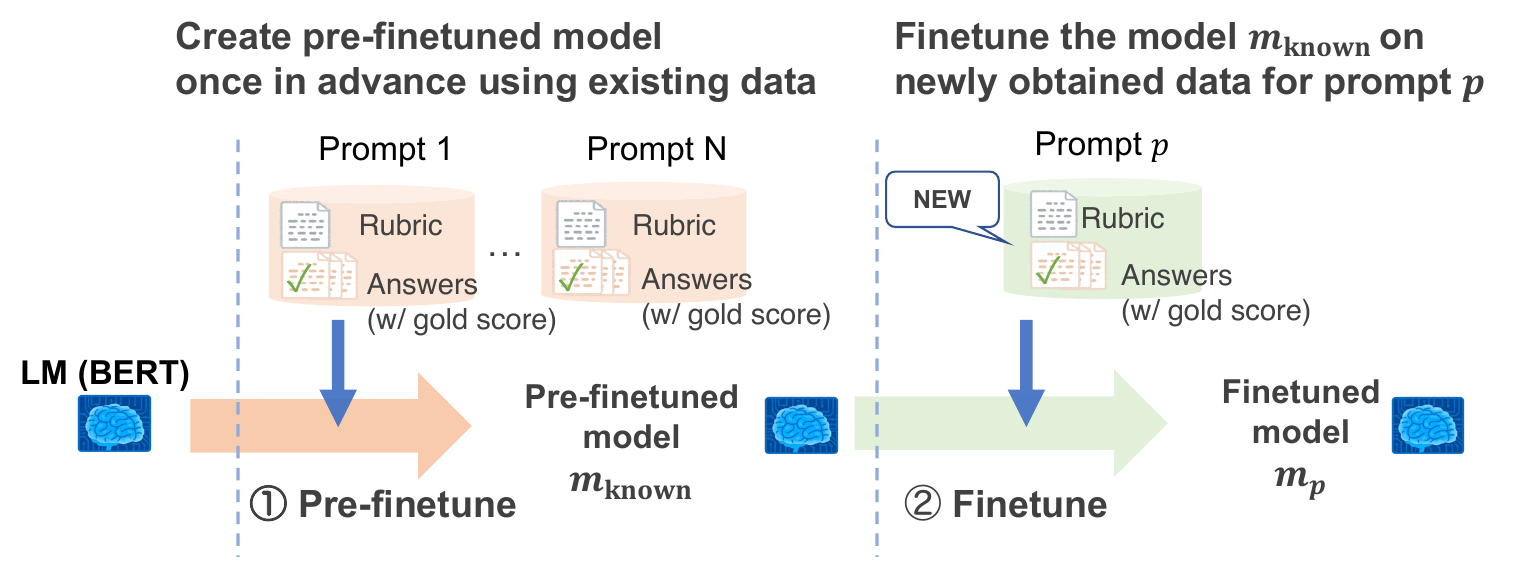}
\caption{Overview of our proposed method. We input key phrases, reference expressions, with an answer. We first pre-finetune the SAS model on already annotated prompts and then finetune the model on a prompt to be graded.}
\label{fig:overview}
\end{figure}

We address both challenges through a new two-phase approach: (i) train (pre-finetune) a model on existing rubrics and answers and (ii) finetune the model for a given new prompt (see Figure~\ref{fig:overview}). 
This approach resolves the data accessibility issue since the second phase (finetuning on a new prompt) does not require access to the cross-prompt data used in the first phase.
Note that the second phase needs access only to the parameters of the pre-finetuned model. 
On the other hand, it is not obvious whether the approach exhibits cross-prompt generalizability. 
However, our experimental results indicate that a SAS model can leverage cross-prompt training data to boost the scoring performance if it is designed to learn the task's property shared across different prompts effectively.

Our contributions are as follows.
(I) Through our two-phase approach to cross-prompt training, we conduct the first study in SAS literature to alleviate the need of expensive training data for training a model on every newly given prompt, while resolving the problem of limited accessibility to proprietary cross-prompt data.
(II) We conduct experiments on a SAS dataset enriched with a large number of prompts (109), rubrics, and answers and show that a SAS model can benefit from cross-prompt training instances, exhibiting a considerable gain in score prediction accuracy on top of in-prompt training, especially in settings with less in-prompt training data.
(III) We conduct an extensive analysis of the model's behavior and find that it is crucial to design the model so that it can learn the task's general property (i.e., a principle of scoring): an answer gets a high score if it contains the information specified by the rubric.

We publicly release our code and all of the experimental settings for reproducing our results at \url{https://ANONYMIZED}

\section{Related work}
\label{sec:related_work}
We position this study as a combination of the use of rubric and domain adoption using cross-prompt data.

To our knowledge, few researchers have focused on using rubrics.
A study~\cite{Wang2021-data-augment-rubrics} used key phrases excerpted from rubrics to generate pseudo-justification cues, a span of the answer that indicates the reason for its score, and they showed the model performance is improved by training attention by pseudo-justification cues.
\cite{sakaguchi-etal-2015-effective} proposed a model that utilizes the similarity between the key concept described in rubrics and the answer.
Following the utilization of rubrics in previous research, we also use the key phrases listed in the rubric, which are typical expressions used for achieving higher scores.

Domain adaptation in this field is also still unexplored.
\cite{sung-etal-2019-pre} further pre-trained BERT~\cite{devlin-etal-2019-bert} on a textbook corpus to learn the knowledge of each subject (i.e. science) and report slight improvement.
On the other hand, we pre-finetune the BERT on cross-prompt data to adopt 
 the scoring task.

Since SAS has different rubrics and reference answers for each prompt, the use of cross-prompt data still remains an open challenge~\cite{Haller-2022-survey-short-answer-deep-learning}. 
As far as we know, the only example is \cite{swarndeep-join-multi-domain-short-answer}. 
However, for each new prompt (i.e., target domain in their term), their model was required to be retrained with both in-prompt and cross-prompt data, which leaves the issue of limited accessibility to proprietary cross-prompt data. 
In contrast, our two-phase approach resolves the data accessibility issue since the second phase (finetuning on a new prompt) does not require access to the cross-prompt data used in the first phase.

\section{Preliminaries}
\begin{figure}[t]
\centering
\includegraphics[width=\textwidth]{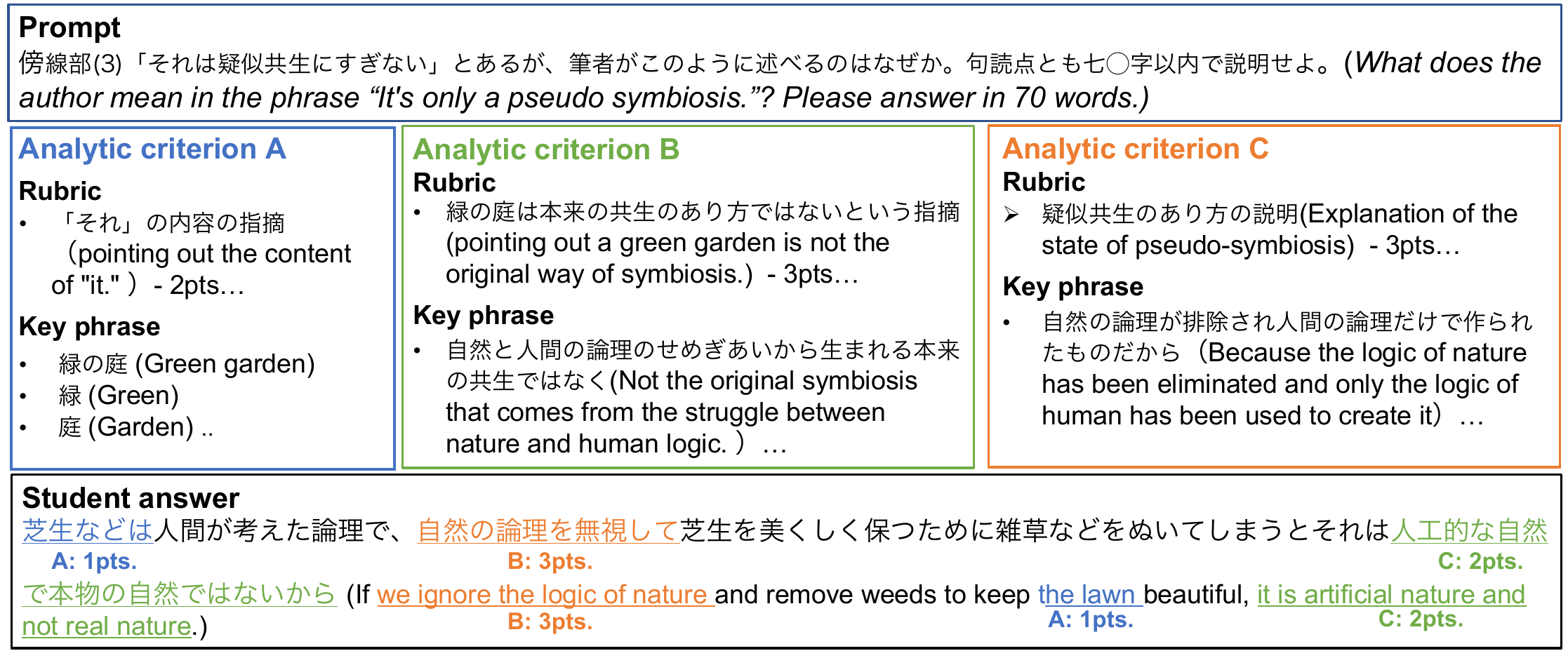}
\caption{Example of a prompt, scoring rubric, key phrase and student's answers excerpted from RIKEN dataset~\cite{mizumoto-etal-2019-analytic} and translated from Japanese to English. For space reasons, some parts of the rubrics and key phrase are omitted.}
\label{fig:rubrics_and_answer}
\end{figure}

\subsection{Task definition}
\label{ssec:task_difnition}

Suppose $X_p$ represents a set of all possible student answers for a prompt $p \in{P}$, and ${\mathbf{x}} \in X_p$ is an answer. 
Each prompt has a discrete integer score range $S = \{0, ..., N_p\}$, which is defined in the rubric.
The score of each answer is chosen within the range $S$.
Therefore, the SAS task is defined as assigning one of the scores $s \in S$ for each given input ${\mathbf{x}} \in X_p$.

In this study, we assume that every prompt is associated with a predefined rubric, which stipulates what information an answer must contain to get a score. 
An answer gets scored high if it contains the required information sufficiently and low if not. 
Figure~\ref{fig:rubrics_and_answer} shows an example of a prompt with a rubric from the dataset we used in our experiments~\cite{mizumoto-etal-2019-analytic}. 
As in the figure, the required information stipulated by a rubric may also be presented by a set of key phrases to help human raters and students understand the evaluation criteria. 
Each key phrase gives an example of wording that gives an answer score high. 
In the dataset used in our experiments, every rubric provides a set of key phrases, and we utilize such key phrases in cross-prompt training. 

In our cross-prompt training setting, we assume that we have some already graded prompts by human raters $P_\mathrm{known}$ and we then want to grade a new prompt $p_\mathrm{target}$ automatically. 
Within this cross-prompt setting, the model is required to score the answers having different score ranges.
Therefore, we re-scale the score ranges of all $P_\mathrm{known}$ and $p_\mathrm{target}$ to $[0, 1]$, and 
as a result, the goal of the task is to construct a regression function $m : \bigcup_{p \in P_\mathrm{known} \cup \{p_\mathrm{target}\}} \{X_p\} \rightarrow [0, 1]$ that maps an student answer to a score $s \in [0,1]$. 

\subsection{Scoring model}
\label{ssec:scoring_model}
A typical approach to construct a function $m$ is to use deep neural networks.
Suppose $\mathcal{D}=( (\mathbf{x}_i, s_i) )^{I}_{i=1}$ is training data that consist of the pairs of an actually obtained student answer $\mathbf{x}_i$ and its corresponding human-annotated score $s_i$. $I$ is the number of training instances.
To train the model $m$, we attempt to minimize the Mean Squared Error loss on the training data $L_{m}(\mathcal{D})$:
\begin{align}
{m^{*}} = \argmin_{m} \left\{ L_{m}(\mathcal{D}) \right\}, \quad 
L_{m}(\mathcal{D})  =
\frac{1}{I}\sum_{(\mathbf{x},s)\in \mathcal{D}} (s - m(\mathbf{x}))^2
,
\end{align}
where $m(\mathbf{x})$ is a score predicted by the model $m$ for a given input $\mathbf{x}$.
Once ${m}^{*}$ is obtained, we can predict the score $s$ of a new
student answer as: $s = m^{*}(\mathbf{x})$.

We construct the model $m$ as following.
Let $\textbf{enc}(\cdot)$ as the encoder, we first obtain a hidden vector $\mathbf{h_x} \in \mathbb{R}^{H}$ from an input 
answer $\mathbf{x}$ as: 
\begin{align}
    \mathbf{h_x} = \texttt{enc}(\mathbf{x}).
\end{align}
Then, we feed the hidden vector $\mathbf{h_x}$ to a linear layer with a sigmoid function to predict a score:
\begin{align}
   m({\mathbf{x}}) = \texttt{sigmoid}(\mathbf{w}^{\top}\mathbf{h_x}+{b}), 
\end{align}
where $\mathbf{w} \in \mathbb{R}^{H}$ and ${b}\in \mathbb{R}$ are learnable parameters.
In this paper, we used BERT~\cite{devlin-etal-2019-bert}, a widely used encoder in various NLP tasks, as the encoder. 


\section{Method}
\label{sec:method}

To leverage cross-prompt training data, we consider the following two-staged training process: (i) We first finetune the model with cross-prompt training instances so that it learns the task's general property (i.e., principles of scoring) shared across prompts, and (ii) we then further finetune the model with in-prompt training instances to obtain the model specific for the target prompt. 
We call the training in the first stage \emph{pre-finetuning}, following~\cite{aghajanyan-etal-2021-muppet}. 
The questions become what kind of general property can the model learn from cross-prompt training instances and how the model learns.

To address these questions, we first hypothesize that one essential property a SAS model can learn in pre-finetuning is the scoring principle: an answer generally gets a high score if it contains sufficient information specified by the rubric and gets a lower score if it contains less. 
The principle generally holds across prompts and is expected to be learned from cross-prompt training instances. 
To learn it, the model needs to have access to the information specified by the rubrics through pre-finetuning and finetuning. 
We elaborate on this below.

\subsubsection{Key phrases}
\begin{figure}[t]
\centering
\includegraphics[width=6cm]{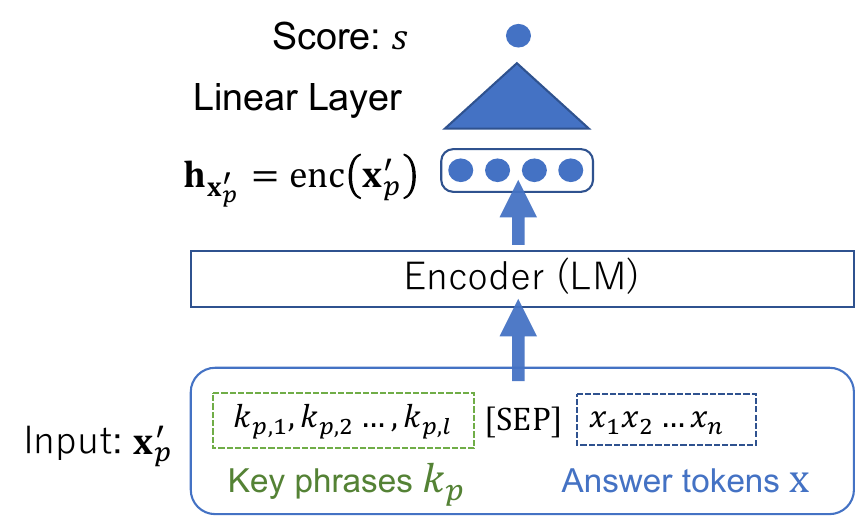}
\caption{Overall architecture of our model. We input key phrases and a student answer split by the [SEP] token.}
\label{fig:overall_architecture}
\end{figure}

As reference expressions for high score answers, we utilize key phrases described in the rubrics as shown in the middle part of Figure~\ref{fig:rubrics_and_answer}.
Key phrases are representative examples of the expressions that an answer must contain in order to gain scores.

Key phrases are clearly stated in each rubrics.
We use those key phrases for each prompt $p$ from the corresponding rubric, and generate a key phrase sequence $k_p$ for $p$ by enumerating multiple key phrases into a single sequence with a comma delimiter.
We then use the concatenated sequence $\mathbf{x}_p^{\prime}$ of tokens $k_p$, {\tt [SEP]}, and $\mathbf{x}$ in this order, as our model input.
For the model without using key phrases, we instead input a prompt ID to distinguish the prompt. We show the overall architecture of the model in Figure~\ref{fig:overall_architecture}.

\subsubsection{Pre-finetuning}
We utilize data from already annotated prompts $P_\mathrm{known}$ to train models for a new prompt $p_\mathrm{target}$.
For each prompt $p \in{ P} $, there exists a key phrase sequence $k_{p}$. We create a concatenated input sequence $\mathbf{x}^{\prime}_{p,i}$ for the $i$-th answer of the prompt $p$ as: $\mathbf{x}^{\prime}_{p,i} = \{k_p, [SEP], \mathbf{x}_{p,i}\}$. Then, we construct data for pre-finetuning as:
\begin{align}
\mathcal{D}_\mathrm{known} = \{(\mathbf{x}^{\prime}_{p,i}, s_{p,i})~|~p \in P_\mathrm{known}\}^{I}_{i=1}.
\end{align}
We pre-finetune the BERT-based regression model on this dataset $\mathcal{D}_\mathrm{known}$ and obtain the model $m_\mathrm{known}$:
\begin{align}
{m_\mathrm{known}} = \argmin_{m} \left\{ L_{m}(\mathcal{D}_\mathrm{known})\right\}.
\end{align}
Next, we further finetune the pre-finetuned model $m_\mathrm{known}$ on $p \in P_\mathrm{target}$ to obtain a model $m_p$ for the prompt $p$.
\begin{align}
{m_{p}} = \argmin_{m} \left\{ L_{m}(\mathcal{D}_p) \right\}
\end{align}

\section{Experiment}

\subsection{Dataset}
\subsubsection{RIKEN dataset}
We use the RIKEN dataset, a publicly available Japanese SAS dataset\footnote{\url{ https://aip-nlu.gitlab.io/resources/sas-japanese}} provided in \cite{mizumoto-etal-2019-analytic}. RIKEN dataset offers a large number of rubrics, prompts, and answers ideal for conducting our experiments.
As mentioned in Section\ref{sec:introduction}, we added 10,000 new data annotations (20 prompts with 500 answers each) to the RIKEN dataset.

RIKEN dataset is a collection of annotated high school students' answers for Japanese Reading comprehension questions.\footnote{Type of question in which the student reads a essay and answers prompts about its content.}
Each prompt in the RIKEN dataset has several scoring rubrics (i.e., analytic criterion~\cite{mizumoto-etal-2019-analytic}), and each answer is manually graded based on each analytic criterion independently (i.e., analytic score). 

In our experiment, we used 6 prompts (21 analytic criterion), same as \cite{mizumoto-etal-2019-analytic}, from RIKEN dataset as $p_\mathrm{target}$ to evaluate the effectiveness of pre-finetuning.
We split answers for these 6 promts as 200 for train data, 50 for dev set and 250 for test set.
For pre-finetuning, we used the remaining 28 prompts (88 analytic criterion), consisting of 480 answers per analytic criterion for training the model and 20 answers per analytic criterion as the dev set.

Following \cite{funayama-2022-balancing}, we treat analytic criterion as an individual scoring task since each analytic score is graded based on each analytic criterion independently.
For simplicity, we refer to each analytic criterion as a single, independent prompt in the experiments.
Thus, we consider a total of 109 analytic criterion as 109 independent prompts in this dataset.

\subsection{Setting}
\label{ssec:setting}
As described in Section~\ref{ssec:scoring_model}, we used pretrained BERT~\cite{devlin-etal-2019-bert} as the encoder for the automatic scoring model and use the vectors of CLS tokens as feature vectors for predicting answers.\footnote{We used pretrained BERT models from  \url{https://github.com/cl-tohoku/bert-japanese} for Japanese.}

\begin{figure}[t]
\centering
\includegraphics[width=\textwidth]{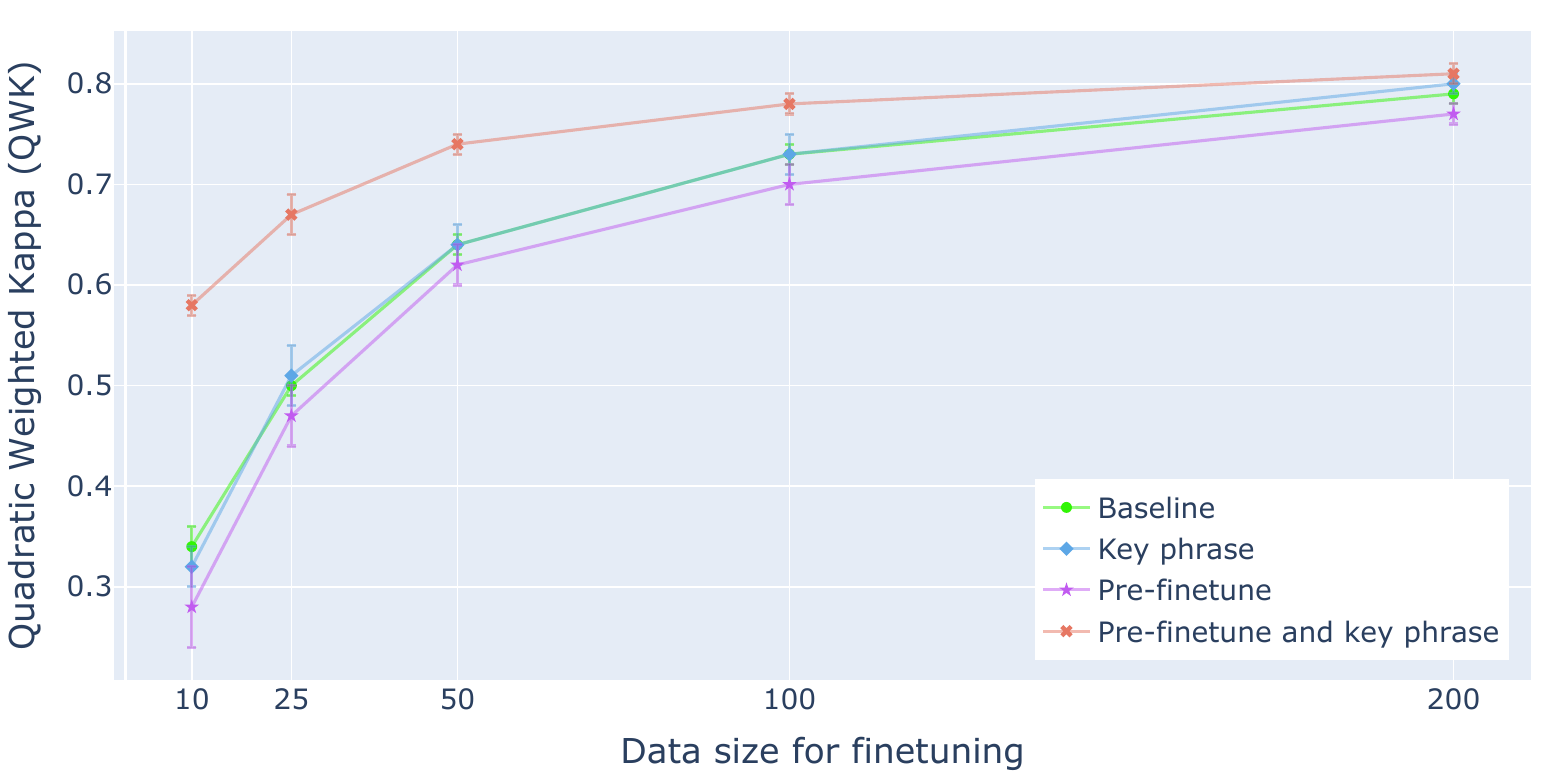}
\caption{QWK and standard deviation of four settings described in Section \ref{ssec:setting}; \texttt{Baseline}, \texttt{Key phrase}, \texttt{Pre-finetune}, and \texttt{Pre-finetune \& key phrase}. In the pre-finetuning phase, we use 88 prompts with 480 answers per prompt. We change the amount of data for finetuning as 10, 25, 50, 100, and 200.}
\label{fig:qwk_changing_datasize}
\end{figure}

Similar to previous studies~\cite{mizumoto-etal-2019-analytic,riordan-etal-2017-investigating,Mohler:11:ACL}, we use Quadratic Weighted Kappa (QWK)~\cite{Cohen:68:Journal}, the de facto standard evaluation metric in SAS, in the evaluation of our models.
The scores were normalized to a range from 0 to 1 according to previous studies~\cite{riordan-etal-2017-investigating,mizumoto-etal-2019-analytic}. 
QWK was measured by re-scaling to the original range when evaluated on the test set.
We train a model for 5 epochs in the pre-finetuning process. We then finetune the resulting model for 10 epochs.
In the setting without pre-finetuning process, we finetune the model for 30 epochs.
These epoch numbers were determined in preliminary experiments with dev set.
During the finetuning process, we computed the QWK of the dev set at the end of each epoch and stored the best parameters with the maximum QWK.

To verify the effectiveness of cross-prompt pre-finetuning, we compare the following four settings in experiments; \texttt{Baseline}: Only finetune the BERT-based regression model for a target prompt without key phrases, the most straightforward way to construct a BERT-based SAS model. \texttt{Key phrase}: Only finetune BERT for a target prompt, we input an answer and key phrases to the model. \texttt{Pre-finetune}: Pre-finetune BERT on cross-prompt data, input only an answer. \texttt{Pre-finetune \& key phrase}: Pre-finetune BERT on cross-prompt data, input an answer and key phrase pairs.

\subsection{Results}

First, to validate the effectiveness of the pre-finetuning with key phrases, we examined the performance of the models for the four settings described in Section \ref{ssec:setting}.
Here, similar to~\cite{mizumoto-etal-2019-analytic}, we experimented with 10, 25, 50, 100, and 200 training instances in the finetuning phase.
The results are shown in Fig.~\ref{fig:qwk_changing_datasize}.
We can see that pre-finetune without key phrases slightly lowers the model performance compared to Baseline.
As expected, this result indicates that simply pre-finetuning on other prompts is not effective.
Similarly, using only key phrases without pre-finetuning does not improve performance.
QWK improves significantly only when key phrases are used and when pre-finetune is performed.
The gain was notably large when the training data was scarce, with a maximum improvement of about 0.25 in QWK when using 10 answers for finetuning compared to the Baseline.
Furthermore, our results indicate that the pre-finetuning with key phrases can reduce the required training data by half while maintaining the same performance.
On the other hand, the performance did not improve when we used 200 answers in training, which indicates that pre-finetuning does not benefit when sufficient training data is available.
We note that the results of baseline models are comparable to the results of the baseline model shown in \cite{mizumoto-etal-2019-analytic}.

\subsubsection{Impact of the number of prompts used for pre-finetuning}
\begin{figure}[t]
\centering
\includegraphics[width=\textwidth]{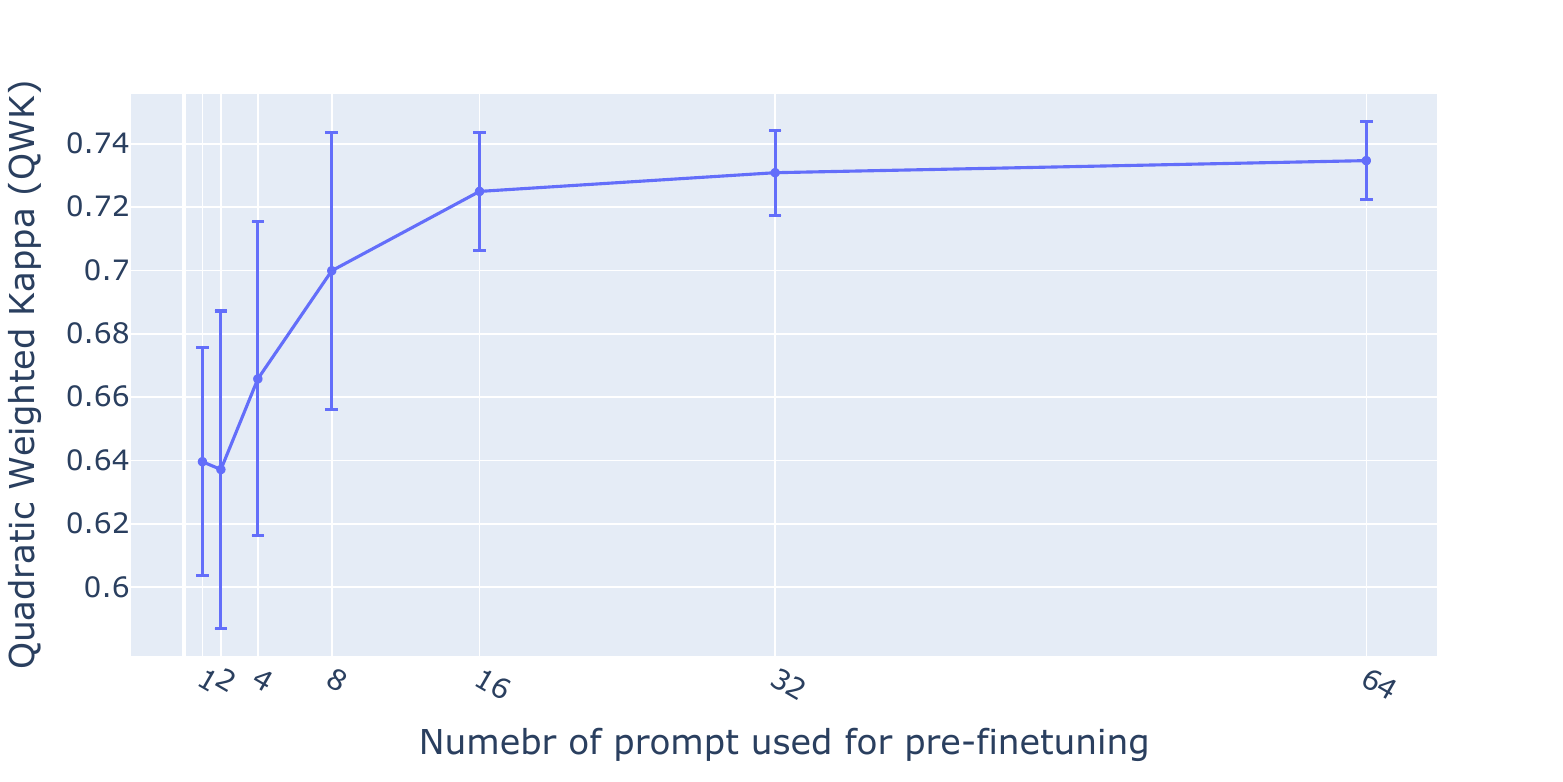}
\caption{QWK and standard deviation when the total number of answers used for pre-finetuning is fixed at 1,600 and the number of prompts used is varied from 1, 2, 4, 8, 16, 32, 64. For finetuning, 50 training instances were used.}
\label{fig:qwk_changing_prompt_num}
\end{figure}

Next, we examined how changes in the number of prompts affect pre-finetuning: we fixed the total number of answers used for the pre-finetuning at 1,600 and varied the number of prompts between 1, 2, 4, 8, 16, 32, 64.
We performed finetuning using 50 answers for each prompt.
The results are shown in Table~\ref{fig:qwk_changing_prompt_num}.
We see that the performance increases as the number of prompts used for pre-finetuning is increased.
This result suggests that the more diverse the answer and key phrases pairs, the better the model understands their relationship.
It also suggests that increasing the number of prompts is more effective for pre-finetuning than increasing the number of answers per prompts.

We can see the large standard deviation when the number of prompts used for pre-finetuning is small.
We assume that the difference in the sampled prompts caused the large standard deviation, suggesting that some prompts might be effective for pre-finetuning while others are unsuitable for pre-finetuning.
The result suggests that a certain number of prompts is needed for training in order to consistently obtain the benefits of cross-prompt learning for each new prompt. We also need some evaluation method for generality of the obtained pre-finetuned model, such as cross-validation among the training prompts.

\subsection{Analysis: what does the SAS model learn from pre-finetuning on cross prompt data?}

We analyzed the behavior of the model in a zero-shot setting to verify what the model learned from pre-finetuning on cross-prompt data.

First, we examined the performance of the \texttt{Pre-finetune \& key phrase} model in a zero-shot setting. 
As a result, we observed higher QWK scores for some prompts, as 0.81 and 079 points in the best-performing two prompts Y14\_2-2\_2\_3-B and Y14\_1-2\_1\_3-D, respectively.
The results indicate that the model somewhat learns the scoring principle in our dataset through pre-finetune using the key phrases; i.e., an answer generally gets a high score if it contains sufficient information specified by the input key phrases.

\begin{figure}[!ht]
\centering
\includegraphics[width=\textwidth]{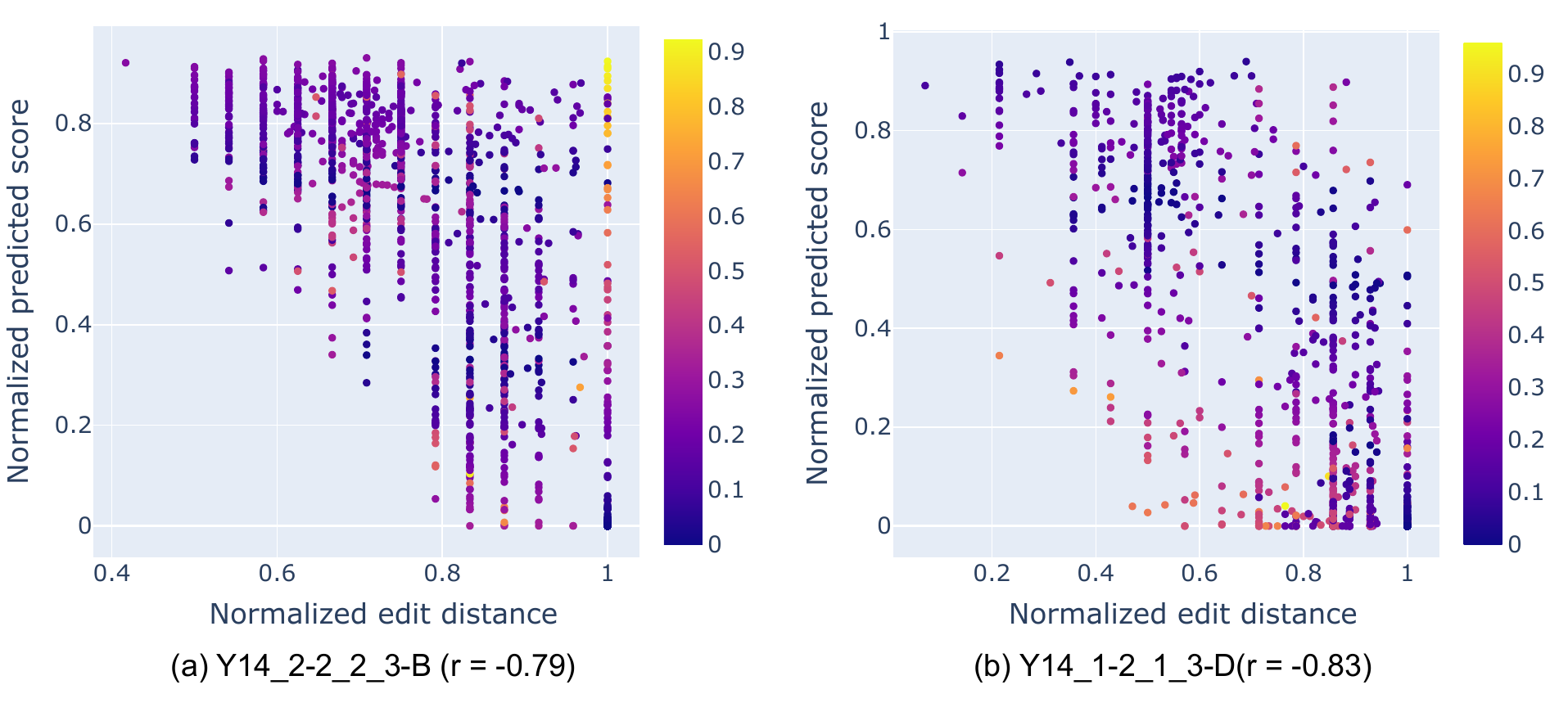}
\caption{Relationship between (x-axis) the normalized edit distance between the justification cue and key phrases in each answer to (a) Y14\_2-2\_2\_3-B and (b) Y14\_1-2\_1\_3-D and (y-axis) the predicted score in zero-shot settings. 
The color bars represent the absolute error between a predicted score and a gold score. $r$ indicates the correlation coefficient.
}
\label{fig:zeroshot_coref}
\end{figure}

Next, to examine how the key phrases contribute to the scoring, using the above two best-performing prompts,
we examined the similarity between the the key phrases and the manually annotated justification cues~\cite{mizumoto-etal-2019-analytic} (substrings of an answer which contributed to gain the score) in the student answer.
For the similarity measure, we employ the normalized edit distance.
Then we analyze the relationship between the edit distance and the predicted scores by the model.

The results for the two prompts with the highest QWK, Y14\_2-2\_2\_3-B, Y14\_1-2\_1\_3-D, are shown in Figure~\ref{fig:zeroshot_coref}.
The color bars represent absolute error between a predicted score and a gold score.
The correlation coefficients are -0.79 and -0.83, respectively, indicating a strong negative correlation between edit distance and predicted scores.
This suggests that the more superficially distant the key phrases and answer, the lower the predicted model score will be for an answer.
We also see that the model correctly predicts a variety of score points for the same edit distance values. 
We show some examples that have lower prediction error with high edit distance in Table~\ref{tb:example_zero_shot}. The examples indicate that the model predicts higher scores for answers that contain expressions that are semantically close to key phrases.

Those analysis indicate that the model partially grasp the property of the scoring task, in which an answer gains higher scores if the answer includes an expression semantically closer to the key phrases.
Such a feature could contribute to the model's high performance, even when the model could not learn enough answer expression patterns from small training data.
\begin{table}[t]
\centering
\caption{Examples of key phrases, answers, predicted scores (Pred.), normalized human annotated scores (Gold.), and normalized edit distance (Dist.). Examples are excerpted from the prompts (1) Y14\_2-2\_2\_3-B and (2) Y14\_1-2\_1\_3-D. Sentences are partially omitted due to the space limitation.}
\begin{tabular}{@{}llcc@{}}
\toprule
Key phrases               & Answers       & Pred. & Gold. \\ \midrule
 \begin{tabular}{l}
(1) 真実よりも幸福を優先する \\(Prioritize happiness over truth..)
  \end{tabular}& \begin{tabular}{l}幸福のためにはどうすれば良いか\\ということについてばかり考える\\(.. only think about how to \\realize happiness.) \end{tabular}&   0.36  & 0.33  \\ 
     \rowcolor{LightGray} \begin{tabular}{l}
(2) 言葉を尽くして他人を説得する\\(Convince others with all my words)
  \end{tabular} & \begin{tabular}{l}説得に努まなければならない.. \\(..to try hard to convince others.) \end{tabular} & 0.50     & 0.50\\
  
\bottomrule
\end{tabular}
\label{tb:example_zero_shot}
\end{table}

\section{Conclusion}

In SAS, answers for each single prompt need to be annotated in order to construct a highly-effective SAS model specifically for that prompt.
Such costly annotations are a major obstacle in deploying SAS systems into school education and e-learning courses, where resources are extremely limited.
To alleviate this problem, we introduced a two-phase approach: train a model on cross-prompt data and finetune it on a new prompt.
Given that scoring rubrics and reference answers are different in every single prompt, we cannot use them directly to train the model.
Therefore, we utilized key phrases, or representative expression that answer should contain to gain scores, and pre-finetune the model to learn the relationship between key phrases and answers.

Our experimental results showed that pre-finetuning with key phrases greatly improves the performance of the model, especially when the training data is scarce (0.24 QWK improvement over the baseline for 10 training instances).
Our results also showed that pre-finetuning can reduce the amount of required training data by half while maintaining similar performance.
As our analysis showed, mere domain adoption by pre-finetuning on cross prompt data is not effective, and it is essential to train the model in terms of the relationship between key phrases and answers to benefit pre-finetuning.

\subsubsection{Acknowledgments}
We are grateful to Dr. Paul Reisert for their writing and editing assistance.
This work was supported by JSPS KAKENHI Grant Number 22H00524, JP19K12112, JST SPRING, Grant Number JPMJSP2114.
We also thank Takamiya Gakuen Yoyogi Seminar for providing invaluable data useful for our experiments.

\bibliographystyle{splncs04} 
\bibliography{references}
\end{document}